\def\year{2022}\relax
\documentclass[letterpaper]{article} 
\usepackage{aaai22}  
\usepackage{times}  
\usepackage{helvet}  
\usepackage{courier}  
\usepackage[hyphens]{url}  
\usepackage{graphicx} 
\usepackage{natbib}  
\usepackage{caption} 
\DeclareCaptionStyle{ruled}{labelfont=normalfont,labelsep=colon,strut=off} 
\usepackage{algorithm}
\usepackage{algorithmic}

\usepackage{newfloat}
\usepackage{listings}
\floatstyle{ruled}
\newfloat{listing}{tb}{lst}{}
\floatname{listing}{Listing}

\usepackage{color}

\definecolor{myPurple}{rgb}{0.4, .0, .8}
\definecolor{myGreen}{rgb}{0, .8, .3}
\definecolor{myRed}{rgb}{0.8, .2, .2}
\definecolor{myOrange}{rgb}{0.8, 0.45, 0.0}
\definecolor{myBlue}{rgb}{.0, .0, 1.0}

\title{Detailed Facial Geometry Recovery from Multi-View Images \\by Learning an Implicit Function}
\author {
    Yunze Xiao\textsuperscript{\rm 1}$^*$,
    Hao Zhu\textsuperscript{\rm 1}$^*$,
    Haotian Yang\textsuperscript{\rm 1},
    Zhengyu Diao\textsuperscript{\rm 1},
    Xiangju Lu\textsuperscript{\rm 2},
    Xun Cao\textsuperscript{\rm 1}$^\dag$
}

\affiliations {
    \textsuperscript{\rm 1} Nanjing University, Nanjing, China\\
    \textsuperscript{\rm 2} iQIYI Inc, Beijing, China\\
    \{yunze.xiao, diaozhengyu\}@smail.nju.edu.cn, \{zhuhaoese, caoxun\}@nju.edu.cn, \\
    yanght321@gmail.com, luxiangju@qiyi.com
}

\usepackage{amsmath}
\usepackage{bm}
\def\etal{\emph{et al.}}

\begin{document}

\maketitle

\let\thefootnote\relax\footnotetext{\hspace{-0.22in}$^*$ These authors contributed equally to this work.\\ $^\dag$ Xun Cao is the corresponding author. }

\begin{abstract}
Recovering detailed facial geometry from a set of calibrated multi-view images is valuable for its wide range of applications. Traditional multi-view stereo (MVS) methods adopt an optimization-based scheme to regularize the matching cost. Recently, learning-based methods integrate all these into an end-to-end neural network and show superiority of efficiency. In this paper, we propose a novel architecture to recover extremely detailed 3D faces within dozens of seconds.  Unlike previous learning-based methods that regularize the cost volume via 3D CNN, we propose to learn an implicit function for regressing the matching cost.  By fitting a 3D morphable model from multi-view images, the features of multiple images are extracted and aggregated in the mesh-attached UV space, which makes the implicit function more effective in recovering detailed facial shape. Our method outperforms SOTA learning-based MVS in accuracy by a large margin on the FaceScape dataset. The code and data are released in \emph{https://github.com/zhuhao-nju/mvfr}.
\end{abstract}

\section{Introduction}
Recovering high-quality facial 3D models is of great value in many fields including the movie industry, facial analysis, and augmented reality.  Multi-view stereo (MVS) focuses on reconstructing accurate shape from several calibrated images, which is the main way to obtain high-quality 3D face models.  
Traditional MVS pipeline mainly consists of matching cost computation, fusion, and refinement\cite{beeler2010high-quality, furukawa2010accurate, galliani2015massively}, which can yield extremely fine facial geometry at the cost of tremendous calculations and running time.
In recent years, learning-based MVS methods\cite{chen2019point-based, gu2020cascade, huang2018deepmvs, im2019dpsnet, ji2017surfacenet, kar2017learning, luo2019p-mvsnet, yao2018mvsnet, yao2019recurrent} show superiority in accuracy and speed, which use a single-pass neural network instead of traditional iterative optimization.  Though learning-based MVS outperforms traditional methods in certain scenes, there is still no learning-based MVS method that recovers fine-level facial geometry for industrial 3D modeling. 

\begin{figure}[t]
\begin{center}
    \includegraphics[width=1.0\linewidth]{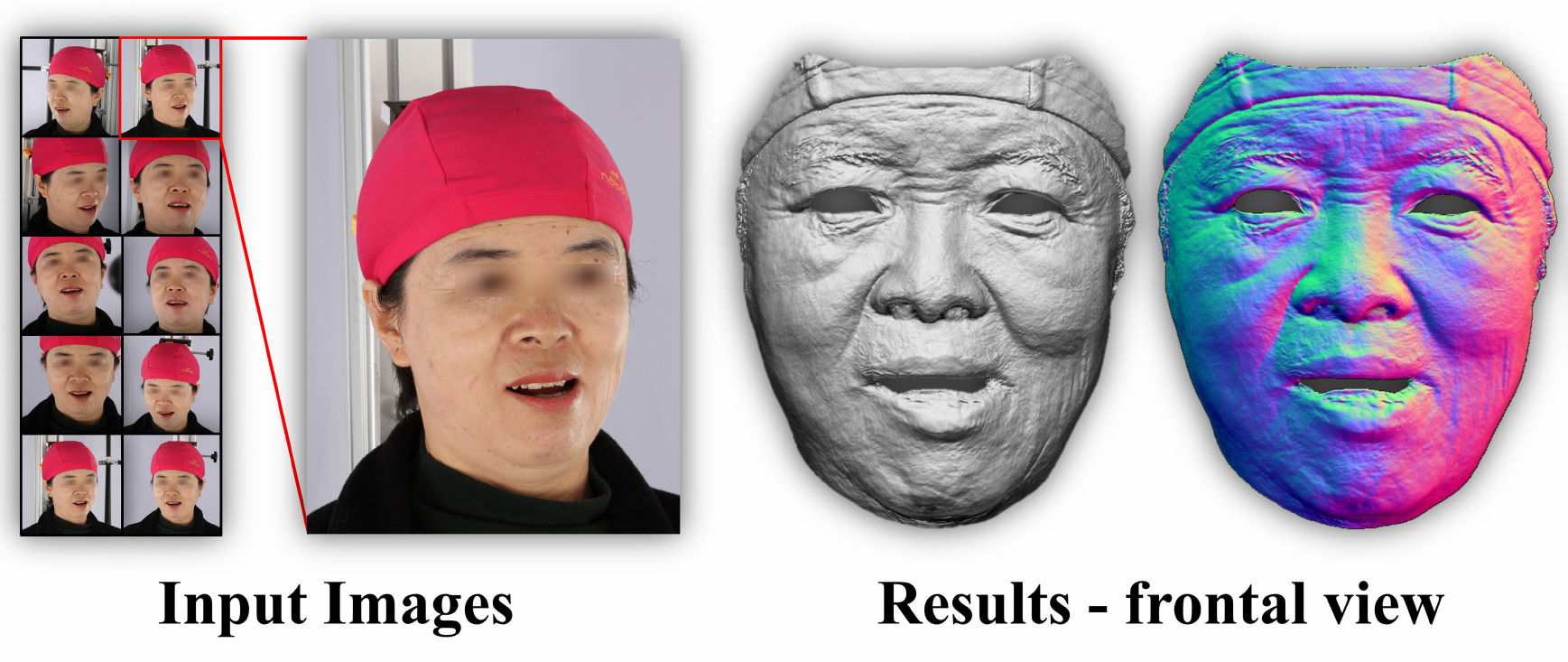}
\end{center}
    \caption{Our method reconstructs detailed facial geometry from multi-view images. The result shape is rendered in grey (left) and with normal-color (right). The input image is partially mosaicked for anonymization.}
\label{fig:teaser}
\end{figure}

In this paper, we propose a novel framework to efficiently recover detailed facial geometry from calibrated multi-view images. The key observation is that the previous state-of-the-art learning-based methods\cite{yao2018mvsnet, yao2019recurrent, chen2019point-based, gu2020cascade, yi2020PVAMVSNET, yan2020dense} build and refine the cost volume in the camera frustum space, which is memory-consuming and limits the resolution.  Though several works\cite{gu2020cascade, riegler2017octnet, wang2017o-cnn} tried to improve the efficiency of the volume-based neural network, they didn't utilize the facial prior and fail to recover the detailed geometry.  To overcome this problem, we propose to use an implicit function to regularize the matching cost to recover geometric details.  To make the framework achievable, multi-view 3DMM estimation is utilized to transfer the cost space into UVD space (UV + displacement), which efficiently solves the rough shape, and supports the implicit function learning in detailed shape recovery.  Compared to the previous learning-based MVS framework, our method doesn't require an additional fusion phase, which saves processing time and achieves higher accuracy. 

The contribution can be summarized in three aspects:
\begin{itemize}

\item A novel facial-specific multi-view reconstruction pipeline is proposed consisting of base model fitting, learned implicit function, and mesoscopic prediction.

\item We propose to tackle the problem of cost regression and multi-view fusion simultaneously by learning an implicit function in our network, which is proved to be more accurate and time-saving.

\item We propose to build the mesh-attached cost volume in UVD space, which greatly reduces the solution space for implicit function learning, achieving efficient and effective MVS for human faces. 

\end{itemize}

\section{Related Works}
The related works can be divided into three categories: traditional MVS, learning-based MVS, and face-specific MVS.

\textbf{Traditional MVS.}  Traditional MVS recovers the 3D shape of the objects or scenes from a set of calibrated multi-view images.  
According to the representation of the 3D modeling, traditional MVS can be categorized into volumetric method\cite{kutulakos2000a, seitz1997photorealistic}, point-based method\cite{furukawa2010accurate, lhuillier2005a, chen2019point-based} and depth-based method\cite{campbell2008using, galliani2015massively, schonberger2016pixelwise, tola2012efficient, liu2009continuous}.  Due to the space limitation, we recommend referring to the survey/benchmark\cite{seitz2006a, knapitsch2017tanks, zhu2017role} and the tutorial\cite{furukawa2015multi-view} for the comprehensive review for traditional MVS. Here we focus on reviewing recently proposed learning-based MVS and face-specific MVS, which are more relevant to our work.

\textbf{Learning-based MVS.}  Learning-based MVS methods seek to improve the reconstructing performance by introducing deep neural networks.  In early attempts, Hartmann \etal \cite{hartmann2017learned} proposed to use the encoder network with multiple Siamese input branches to measure the similarity of the image patches from the multi-view input images.  Ji \etal \cite{ji2017surfacenet} proposed to use a CNN to predict each voxel a binary attribute indicating whether the voxel is on the surface.  Kar \etal \cite{kar2017learning} proposed an end-to-end learning MVS network by involving differentiable feature projection and inverse-projection along viewing rays.  
Based on this framework, Yao \etal~\cite{yao2018mvsnet, yao2019recurrent} built the cost volume on the camera frustum, and adopts 3D CNN or RNN to regularize the matching cost.  
Huang \etal~\cite{huang2018deepmvs} proposed to use the CNN to estimate the disparity from the plane-sweep volumes generated from multi-view images.  Sunghoon~\etal \cite{im2019dpsnet} introduced the cost fusion and aggregation network, and finally regress a depth map from the refined cost volume. Chen \etal \cite{chen2019point-based} proposed to process the target scene as point clouds in a neural network, which predicts the depth in a coarse-to-fine manner.  Luo \etal~\cite{luo2019p-mvsnet} proposed the patch-based MVS network consisting of a patch-wise aggregation module to generate a matching confidence volume from extracted features, and a hybrid 3D CNN to infer a depth probability distribution. Gu \etal \cite{gu2020cascade} extends the prior volume-based MVS network where cost volume is built upon a feature pyramid encoding geometry and context at gradually finer scales.  Yan \etal~\cite{yan2020dense} proposes a dense hybrid recurrent multi-view stereo net with dynamic consistency checking for accurate dense point cloud reconstruction.  Yi \etal \cite{yi2020PVAMVSNET} presents a pyramid multi-view stereo network with the self-adaptive view aggregation.
Yariv \etal \cite{yariv2020multiview} proposes an implicit differentiable renderer that learns 3D geometry, appearance, and cameras from masked 2D images and noisy camera initialization. Different from all the above methods, our method is the first to learn an implicit function for cost regularization in the learning-based framework.

\begin{figure*}[t]
\begin{center}
    \includegraphics[width=1.0\linewidth]{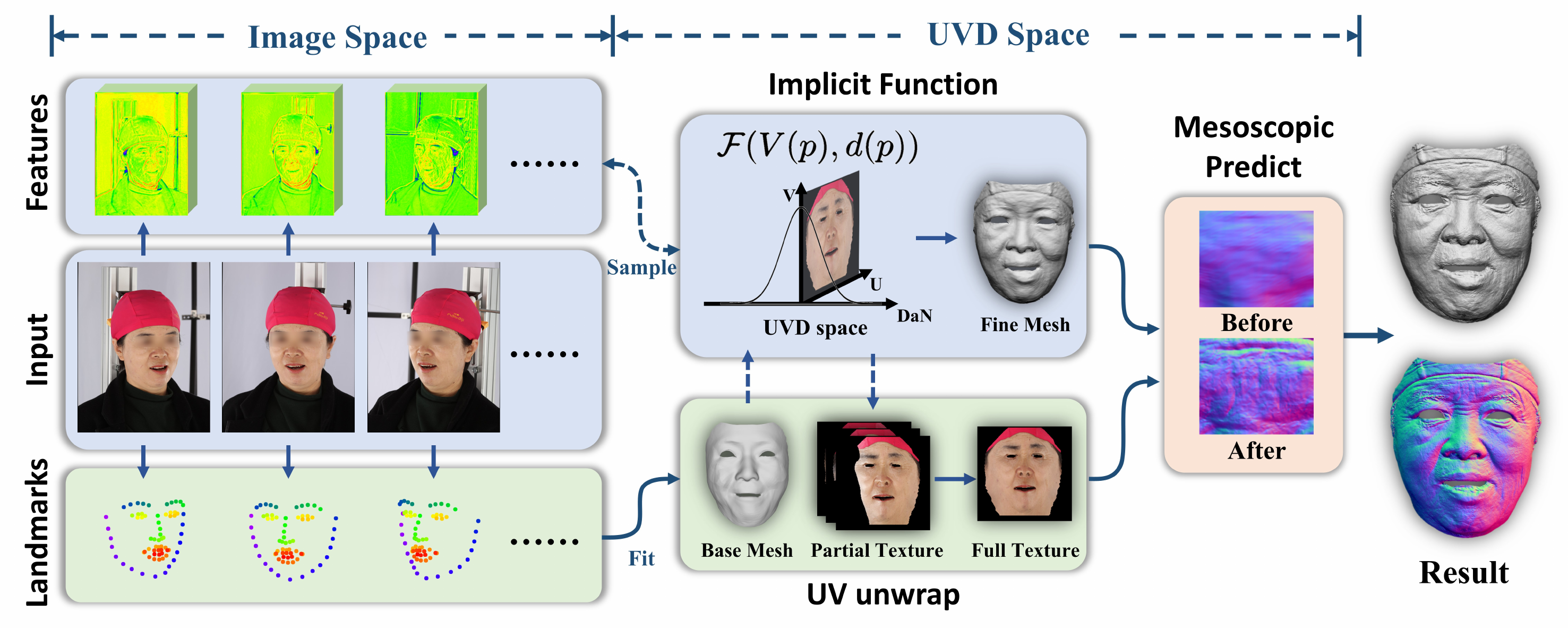}
\end{center}
    \caption{Our pipeline consists of three parts: base mesh fitting (green blocks), implicit function learning (blue blocks), and mesoscopic recovery (pink block).}
\label{fig:pipeline}
\end{figure*}

\textbf{Face-specific MVS.}
Faces are special as they contain relatively fewer photometric features but plenty of geometric details.   Based on the pyramid stereo matching framework, Beeler \etal \cite{beeler2010high-quality} uses smoothness constraint, ordering constraint, and uniqueness constraint to robustly recover the facial shape, then the shape is iteratively optimized to recover pore-level geometry. Bradley \etal \cite{bradley2010high} extends the multi-view facial stereo to multi-view videos that are captured by an array of video cameras. 
Ghosh \etal \cite{ghosh2011multiview} proposes to capture high resolution diffuse and specular photometric information using a multi-view face capture system, then reconstruct detailed facial geometry. Valgaerts \etal \cite{valgaerts2012lightweight} proposed a lightweight passive facial performance capture approach that only requires a single pair of stereo cameras.  
Though generalized MVS methods are able to recover fine 3D models, they consume a lot of computing resources and time to recover faithful geometric details.
In recent years, many works study how to recover non-rigid facial geometry from uncalibrated images or videos.  Dou \etal \cite{dou2018multi} and Ramon \etal \cite{ramon2019multi} proposes to recover the face model from multi-view images using a subspace representation of the 3D facial shape and a deep recurrent neural network to fuse the identity-related features. 
Bai \etal \cite{bai2020deep, bai2021riggable} propose to optimize the 3D face shape by explicitly enforcing multi-view appearance consistency, which makes it possible to recover shape details according to conventional multi-view stereo methods.  

\section{Methods}

As shown in Figure \ref{fig:pipeline}, our method consists of three stages. In the first stage, the 2D landmarks of multi-view images are extracted, and a base mesh is obtained by fitting the 3DMM to these landmarks.  In the second stage, the accurate 3D facial geometry is predicted by learning the implicit function, and the detailed shape is refined by the mesoscopic prediction network in the last stage.  We will explain each module in detail in the following sections.

\subsection{Base Mesh Fitting}

Firstly, we use Bulat \etal's method \cite{bulat2017how} to obtain the 2D landmarks $[l_1, l_2, ..., l_M] \in \mathcal{R}^{M\times2}$ for each image.  Given the camera parameters, the 3D landmarks can be solved by minimizing the energy function below:

\begin{equation}
    E(\mathcal{L}_j) = \sum_{i=0}^N{||\Pi_i(\mathcal{L}_j) - {l}_j||_2}
\end{equation}

\noindent where $i$ is the index of the view, $N$ is the total view number. $\Pi_i$ is the perspective projection of the camera in $ith$ view. $\mathcal{L}_j$ and $l_j$ are the $jth$ 3D and 2D landmarks separately. The $\mathcal{L}_j$ in $[\mathcal{L}_1, \mathcal{L}_2, ..., \mathcal{L}_M] \in \mathcal{R}^{M\times3}$ is solved by least-squares respectively.

After 3D landmarks are obtained, we fit the bilinear model generated by FaceScape\cite{yang2020facescape} to the 3D landmarks and obtain the topologically-uniformed base mesh.  The bilinear model of FaceScape can be formulated as $V = \mathcal{B}(id, exp)$, where $\mathcal{B}$ is a linear mapping from the parameter $id$ and $exp$ to the vertices position $V$ of the facial mesh model.  The model fitting is solved by minimizing the energy function below:

\begin{equation}
    E(\bm{id}, \bm{exp}, \mathcal{T}) = \sum_{j=0}^M||\mathcal{T}(K_j(\mathcal{B}(\bm{id}, \bm{exp}))) - \mathcal{L}_j||_2
\end{equation}
\noindent where $\mathcal{T}()$ is the transformation between two coordinate systems. 
$K_j(\mathcal{M})$ extracts the $jth$ 3D landmarks from the input mesh $\mathcal{M}$ using the predefined indices.  We use the bilinear model generated by Yang \etal \cite{yang2020facescape} in our experiments, and this can be substituted by other 3D facial parametric models. 

\subsection{Mesh-attached Cost Space} 

\begin{figure*}[t]
\begin{center}
    \includegraphics[width=1.0\linewidth]{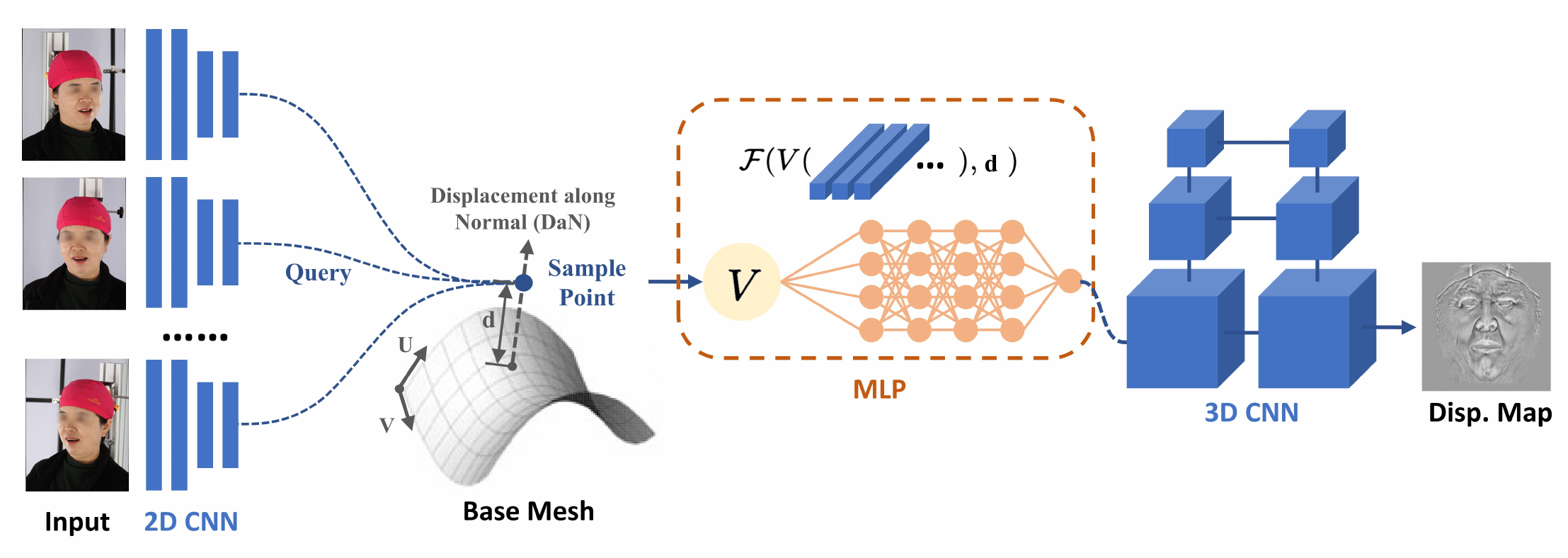}
\end{center}
    \caption{The network in implicit function learning stage. The weights-shared 2D CNN extracts the features of multi-view images. Then the points are sampled in UVD space, and the queried features are fed to a MLP to learn the implicit function. A 3D CNN based post-regularizer is used to extract the displacement map from the predicted cost volume. }
\label{fig:network}
\end{figure*}

Different from the previous method that builds the cost volume in the Euclidean space or camera frustum, our method builds the cost volume in the UVD space attaching to the fitted facial mesh.  The UV map is the flat 2D representation of the surface of the 3D mesh, where the attributes that are attached to the surface can be efficiently stored.  We extend the UV space with the third dimension -- displacement along the surface normal (DaN), and regularize the matching cost in this mesh-attached UVD space. This strategy is similar to the idea of SDF\cite{malladi1995Shape, chan2005level, yariv2021volume} and has been used in many previous 
methods\cite{zhu2019detailed, zhu2021detailed, alldieck2019tex2shape}. 
We use the UV mapping defined by FaceScape, and the UV mapping is uniform in all fitted base meshes.
The surface normal is obtained by sampling the vertex normal of the base mesh on the UV map, and the surface normal is smoothed to eliminate messy normal vectors caused by the high-frequency shape.

The establishment of mesh-attached volume brings three benefits to the facial MVS: 1) The mesh-attached volume connects the statistical facial model estimation and the volume-based depth regression method;
2) The solution space to represent geometry is reduced compared to the Euclidean volume\cite{kar2017learning} or camera frustum\cite{yao2018mvsnet, yao2019recurrent, gu2020cascade}, thus the volume resolution can be enhanced;  
3) The mesh-attached space can represent the complete face, while the camera frustum only sees the partial face and requires an additional fusion phase to produce complete facial geometry.

\subsection{Implicit Function Learning}
\textbf{Implicit function.} Inspired by the success of implicit function in single-view human shape recovery\cite{saito2019pifu, saito2020pifuhd}, we integrate implicit function learning into facial multi-view stereo. The overall network architecture is shown in Figure~\ref{fig:network}. Our implicit function $\mathcal{F}$ is formulated as:

\begin{equation}
    \mathcal{F}(V(p), d(p)) = s, s \in \mathcal{R}
    \label{equ:if}
\end{equation}

where $d(p)$ is the displacement corresponding to the sampled point $p$.  $V(p)$ is the variance of the pixel-wise features extracted from multi-view images, formulated as:

\begin{equation}
    V(p) = \frac{1}{N}\sum\limits_{i=0}^{N}{F_i^c(\Pi_i(p)})^2 - (\frac{1}{N}\sum\limits_{i=0}^{N} F_i^c(\Pi_i (p)))^2
\end{equation}

where $F_i^c$ is the feature with $c$ channels extracted from $ith$ image using the feature extracting network; $\Pi_i$ is the projection function of the $ith$ camera; $N$ is the number of the input images.  

We use the trainable multi-layer perceptron (MLP) to learn this implicit function. The label $\hat{s}$ for training is the distance-to-surface, which is the normalized probability to describe how close from the sampled point to the ground-truth surface. The label is formulated as:

\begin{equation}
    \hat{s}(d) = \frac{1}{\sqrt{2\pi}\sigma_1}e^{-\frac{(d-\hat{d})^{2}}{2{\sigma_1}^{2}}}
\end{equation}

where $\sigma_1$ is set to $0.05$ in our experiment.  Our distance-to-surface label is different from the inside/outside label used in PIFu\cite{saito2019pifu, saito2020pifuhd}, and we will demonstrate the effectiveness in the ablation study section. 

\textbf{Feature extractor.} We use a convolutional neural network to extract the features $F^c_i$ in Equation \ref{equ:if} from the input images. The architecture of our feature extractor is modified based on the feature extractor of MVSNet\cite{yao2018mvsnet} and RMVSNet\cite{yao2019recurrent}.  In our feature extractor, the group normalization\cite{wu2018group} is integrated into each convolutional layer, and half of the striding operations are removed to enhance the resolution of the generated feature maps.  The feature extractor and the implicit function are trained in an end-to-end manner. 

\textbf{Surface-aware sampling.} Compared to the CNN-based regularization network in the previous works\cite{yao2018mvsnet, yao2019recurrent, gu2020cascade, yi2020PVAMVSNET, yan2020dense}, the implicit function benefits from the surface-aware sampling strategy, which uses more points close to the surface for training.  This strategy has been used for human shape recovery in the 3D volumetric space\cite{saito2019pifu, saito2020pifuhd}, while we extend it to samples in the UVD space with a Gaussian distribution. Specifically, the probability distribution to sample in the UVD space follows Gaussian distribution in displacement dimension, and uniform distribution in U and V dimensions. The probability is formulated as:

\begin{equation}
    P(u, v, d) = \frac{1}{\sqrt{2\pi} A }e^{-\frac{(d-\hat{d})^{2}}{2{\sigma_2}^{2}}}
\end{equation}

where $u$, $v$, $d$ are the coordinates along U, V, and displacement axes. $\hat{d}$ is the ground-truth displacement. $A$ is the constant to normalize the integral value of probability to 1.  In our experiments, we set $\sigma_2=0.1$ and the range of $d$ is normalized to $[-1, 1]$. In order to sample the points far away from the surface, an additional uniform sampling in mesh-attached UVD space is adopted. We combine surface-aware sampled points and uniformly sampled points with a ratio of $16:1$.

\begin{figure*}[t]
\begin{center}
    \includegraphics[width=0.875\linewidth]{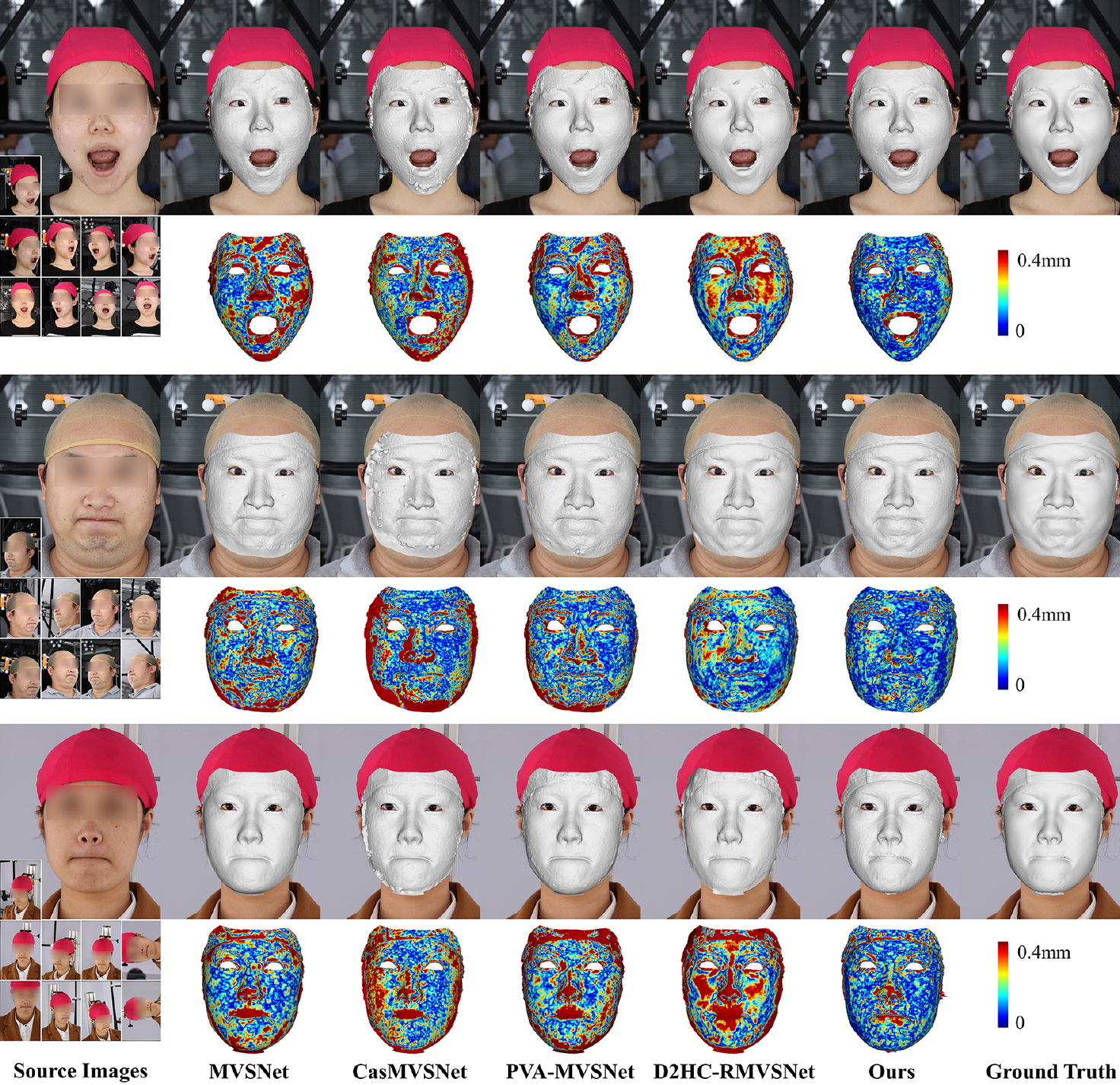}
\end{center}
    \caption{The rendered results and heat maps of ours and previous methods on FaceScape dataset.}
\label{fig:vis_comp}
\end{figure*}

\textbf{Post-regularization.} In the testing phase, the points in the complete UVD space are fed to the feature extractor and the implicit function, obtaining the cost volume in UVD space. A 3D convolutional neural network is used to further refine the cost volume, and the displacement map is extracted from the regularized cost volume using $soft argmax$\cite{kendall2017end} along the displacement dimension.  Here we use $soft argmax$ as it is the differentiable substitution of $argmax$.  The predicted displacement map is applied to the base mesh, generating the final mesh. 

\subsection{Mesoscopic Prediction}
The method in the last Section can recover fine middle-scale geometry, but still lacks the micro-scale geometry such as skin pores and minor wrinkles, because the disparity change across such a feature is too small to detect.  Inspired by the traditional MVS method\cite{beeler2010high-quality}, we notice the fact that light reflected by a diffuse surface is related to the integral of the incoming light, where the small concavities may block part of the incoming light and appear darker.  This fact has been adopted in prior works\cite{beeler2010high-quality, glencross2008perceptually} to synthesize the bump based on the rough mesh. Here we seek the potential to recover a displacement map representing the mesoscopic geometry from the multi-view images.

To be specific, we firstly use the resulting mesh to transfer the texture from the multiple images into the UV space, forming the partial UV textures. For each pixel in the UV texture, we compute the cosine of the angle between the normal direction and the camera direction as the weight, then blend these multi-view UV textures into a complete UV texture using the weighting average.  The blended UV texture is fed to the mesoscopic network to predict the mesoscopic displacement map. The mesoscopic network uses the Pix2PixHD\cite{wang2018pix2pixHD} network as the backbone and is trained using the displacement maps provided by FaceScape dataset. 

\section{Experiments}

\begin{table}[t]
\begin{tabular}{lccc}
\hline
Method                & CD-mean/$mm$ & CD-rms/$mm$ & Comp./$\%$ \\ \hline
MVSNet  & 0.378 / 0.305    & 0.333 / 0.308   & 74.8 / 87.2    \\
CasMVS  & 0.381 / 0.319    & 0.323 / 0.282   & 80.8 / 90.7    \\
PVA  & 0.409 / 0.323    & 0.347 / 0.295   & 77.4 / 92.2    \\
D2HC  & 0.382     & 0.324   & 83.5    \\
Ours                  & \textbf{0.175}         & \textbf{0.202}        & \textbf{100.0}         \\ \hline
\end{tabular}
* Numbers before and after `/' represent the original trained model and the fine-tuned model respectively.
\caption{Comparison with Model-Generalized Methods}
\label{tab:quantitative}
\end{table}

\subsection{Implementation Details}
\label{sec:implementation}
\textbf{Data Preparation.} We use FaceScape\cite{yang2020facescape, zhu2021facescape} dataset to train and validate our method. The FaceScape dataset contains 7120 multi-view images and corresponding 3D models, each of which consists of roughly 50 images, corresponding calibrations, and the 3D mesh model.  These data are captured from 359 different subjects and each in 20 expressions.  We manually filter out the tuples with blur or calibration problems, then selected $80\%$ of the remaining data as the training set and the other $20\%$ as the testing set. For each tuple, we select 10 images from the frontal views as input.  
These images are scaled to $864\times1296$ for training and testing. The authorization of the data is described in the section of Ethics Statement.

In the base mesh fitting phase, we use the bilinear model generated by FaceScape dataset. The officially provided bilinear model is generated from $847$ subjects, which contains the $359$ subjects in our training/testing data. To fairly evaluate our methods, we regenerated the bilinear model with data that excluded the $359$ subjects in our training/test set.  

\textbf{Training Details.} The feature extractor and the implicit function are trained in an end-to-end manner, while the post-regularizer is trained after the other parameters are fixed.  MSE loss is used to train the feature extractor $+$ implicit function, and also the post-regularizer.  We train our network using Adam optimizer, with the learning rate as $10^{-3}$, and our model is trained in $200$ epochs. The batch size is set to $1$.  We trained the model using Nvidia RTX 3090 for about $100$ hours.

\textbf{Metrics.} We use chamfer distance (CD) between the predicted mesh and ground-truth mesh to measure the accuracy of the methods. 
We only measure the accuracy in the facial area, excluding eyes and inner mouth where the ground-truth geometry is hard to obtain.  Considering the result meshes of generalized MVS methods\cite{yao2018mvsnet, gu2020cascade, yi2020PVAMVSNET, yan2020dense} tend to extend out of the facial area, we mask out the outer mesh using the ground-truth mask of the frontal view.  We run the experiments of comparison and ablation study on our testing sets, which contains $1341$ tuples of data.  For each tuple, there are images in $10$ views, calibration parameters, and the ground-truth facial models.  For all $1341$ tuples of results, we report the mean of the chamfer distance (CD-mean) and the root mean square of chamfer distance (CD-rms). We also report the completeness (Comp.) for other methods, which is the area on the ground-truth mesh with error distance  $<2mm$ dividing the area of the full ground-truth mesh. When computing the chamfer distance, the area of the predicted mesh with error distance $>2mm$ is filtered out. Considering our method recovers the complete face, the completeness of our method is 
labeled as $100\%$. All our mesh are counted when computing the chamfer distance.

\begin{figure*}[htbp]
\begin{center}
    \includegraphics[width=0.92\linewidth]{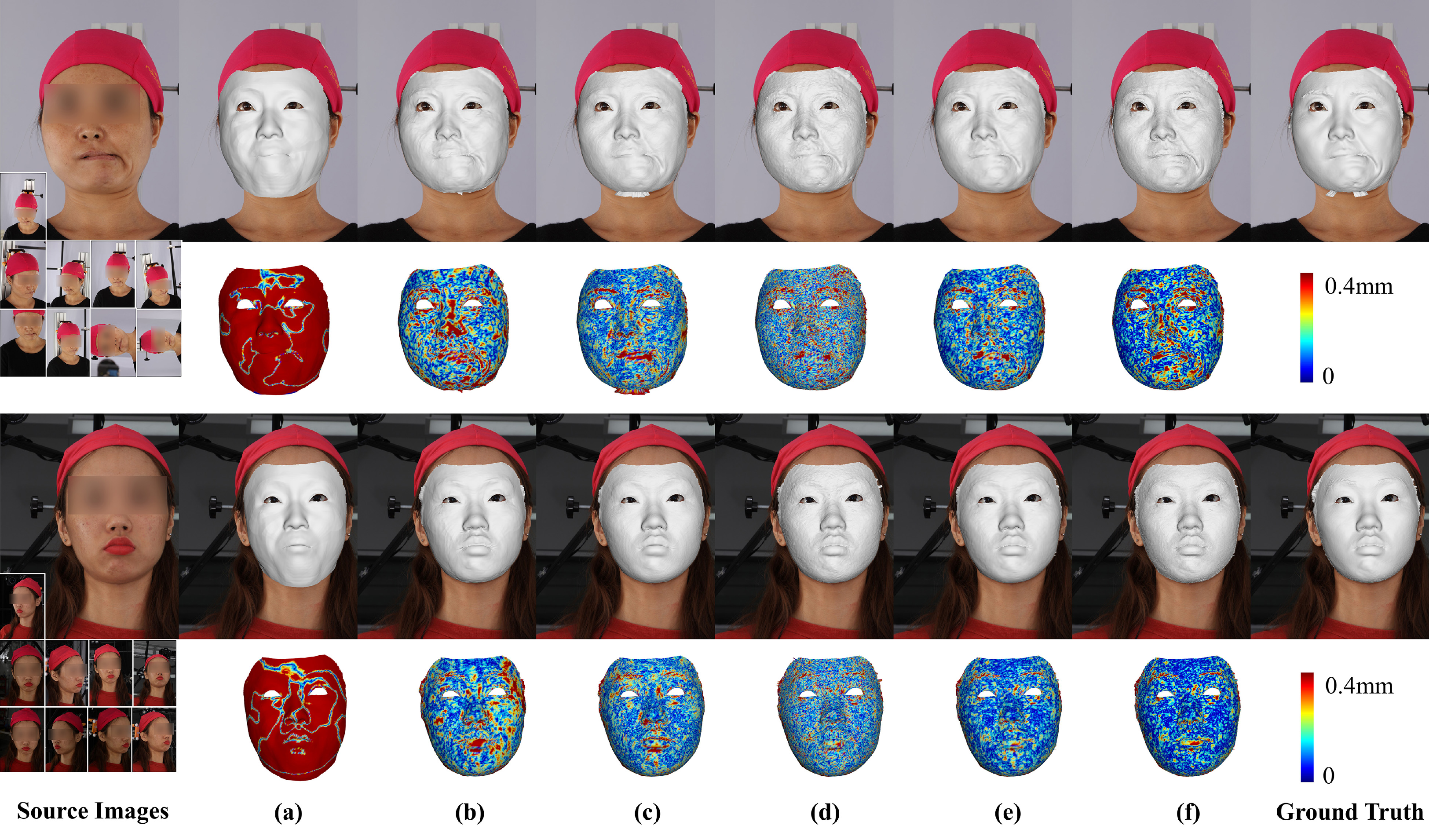}
\end{center}
    \caption{The rendered results and heat maps of the ablation study.}

\label{fig:vis_ablation}
\end{figure*}

\subsection{Comparison with SOTA}

We compare our method with previous methods quantitatively and qualitatively. The baselines are MVSNet\cite{yao2018mvsnet}, CasMVSNet\cite{gu2020cascade}, D2HC-RMVSNet\cite{yan2020dense}, and PVA-MVSNET\cite{yi2020PVAMVSNET}, which are state-of-the-art generalized MVS methods.  We evaluate two models for these methods: `ori' means the originally trained model provided by the authors, which were trained on DTU dataset\cite{aanaes2016large, jensen2014large}; `ft' means the model fine-tuned using our training data generated from FaceScape dataset.  The fine-tuned model is trained with epochs that are equivalent to our training settings.  The scores of CD-mean, CD-rms, and completeness are reported in Table~\ref{tab:quantitative}. The rendered results, as well as heat maps of error distance from predicted mesh to ground-truth mesh, are shown in Figure \ref{fig:vis_comp}, and these results of previous methods are from the fine-tuned models.  

From the quantitative comparison in Table~\ref{tab:quantitative}, we can see that our method outperforms previous methods in CD-mean, CD-rms, and completeness for facial reconstruction.  Among previous methods, though MVSNet\cite{yao2018mvsnet} is slightly better in CD-mean and CD-rms than the other three methods, the completeness of MVSNet is worse, which means more bad areas are not counted in chamfer distance.  We consider the reason is that our method benefits from the mesh-attached UVD cost space and the implicit function learning framework. The enhancement for each proposed module is discussed in the ablation study.

Besides, we compare our method with IDR\cite{yariv2020multiview}, which is a model-specific MVS method that also uses implicit learning for 3D reconstruction. IDR requires to be trained for each object, and the learned parameters are specific to the certain object. By contrast, our method and the methods mentioned above are model-generalized, which means they do not require to be trained for different input images. Since IDR takes an extremely long time to process (about 400 times slower than our method), we only compare the first 5 tuples of multi-view images, and the quantitative results are shown in Table~\ref{tab:comp_idr}. We can see that our method outperforms IDR by a large margin.

\begin{table}[t]
\centering
\begin{tabular}{lccccc}
\hline
Data ID & 1     & 2     & 3     & 4     & 5     \\ \hline
IDR     & 0.971 & 0.744 & 1.129 & 1.065 & 1.003 \\
Ours    & 0.173 & 0.130 & 0.149 & 0.195 & 0.181 \\ \hline
\end{tabular}
\caption{Comparison with Model-Specific Method.}
\label{tab:comp_idr}
\end{table}

\begin{table}[t]
\centering
\begin{tabular}{lccc}
\hline
Method                & CD-mean & CD-rms  \\ \hline
(a) Base                  & 2.821         & 2.987         \\
(b) Base+3DCNN            & 0.230         & 0.269         \\
(c) Base+IF               & 0.195         & 0.212         \\
(d) Base+IF(i/o)+Reg      & 0.181         & 0.222         \\
(e) Base+IF+Reg           & 0.171         & 0.200         \\
(f) Base+IF+Reg+Meso      & 0.175         & 0.202         \\ \hline
\end{tabular}
\caption{Ablation Study (unit:mm)}
\label{tab:ablation}
\end{table}

\subsection{Ablation Study}
\label{sec:ablation}
We evaluate the performance with different components as defined below. For all the settings, the feature extractor and the base mesh fitting phase remain the same.
\begin{itemize}
    \item (a) Base. The base mesh generated by bilinear model fitting. 
    \item (b) Base+3DCNN. The implicit function prediction is replaced by 3D CNN regularizer. This strategy is adopted by prior works\cite{yao2018mvsnet, yao2019recurrent, gu2020cascade}.  
    \item (c) Base+IF. Post-regularize module and mesoscopic prediction are removed.
    \item (d) Base+IF(I/O)+Reg. Mesoscopic prediction is removed, and the implicit function here predicts the inside/outside labels, as used in \cite{saito2019pifu}. 
    \item (e) Base+IF+Reg. Only mesoscopic prediction is removed.
    \item (f) Base+IF+Reg+Meso. Our complete method.
    
\end{itemize}

We visualize the results with different components in Figure~\ref{fig:vis_ablation}, and report the quantitative evaluations of ablation study in Table \ref{tab:ablation}.  
We can see that the base mesh in (a) is quite inaccurate, with chamfer distance $>2mm$.  Comparing (b) with (c), we can see that the implicit function-based framework is more accurate than 3D-CNN-based framework, and the performance of (c) can be further improved by adding a post-regularizer, as shown in (e).  Comparing (d) and (e), we find that distance-to-surface labels are more effective than inside/outside labels.  Comparing (f) with (e), though mesoscopic prediction doesn't enhance the accuracy quantitatively, it improves the visual effects by synthesizing mesoscopic geometry. Comparing (b) with MVSNet in Table \ref{tab:quantitative}, we can see that the cost volume in UVD space outperforms that in camera-frustum, since the much redundant space is ignored in the UVD space. 

The performance for different view numbers are shown in Table~\ref{tab:view_num}. The mean chamfer distance from 10 views to 6 views fluctuates relatively little, but decrease severely when view points are fewer than 5. We believe that this is normal for the multi-view reconstruction method, because fewer viewpoints provide less matching information, resulting in unstable reconstruction.

\begin{table}[]
\centering
\begin{tabular}{lccccc}
\hline
View Num. & 2     & 4     & 6     & 8     & 10    \\ \hline
CD-mean & 5.264 & 0.958 & 0.297 & 0.206 & 0.167 \\
CD-rms  & 5.488 & 1.076 & 0.339 & 0.248 & 0.178 \\ \hline
\end{tabular}
\caption{Test on Different View Numbers (unit:mm)}
\label{tab:view_num}
\end{table}

\section{Conclusion}

In this paper, a novel architecture that leverages parametric model fitting and builds cost volume based on the attached UVD space to recover extremely detailed 3D faces. We propose to learn an implicit function for regularization in the task of multi-view stereo, which is proved to be more efficient and effective in recovering geometric details.

\textbf{Limitation.} There are still unresolved problems with our proposed method. Firstly, our method is only designed for rigid 3D facial reconstruction from well-calibrated and high-quality images. We found that the performance will degrade for the in-the-wild images, and the reason is that the resolution and signal-to-noise ratio of in-the-wild images are commonly too low for extremely fine 3D reconstruction. Secondly, our method can only recover a rigid 3D face, while cannot work for the image sequences with a dynamic face. An expression-identity disentangling module may be further added to solve this problem. Thirdly, it is hard to extend our method to arbitrary objects other than human faces, as our implicit function is built on the base of a parametric model.


\section{Ethics Statement}

The proposed method will promote the development of 3D facial modeling technology, which can be useful for applications requiring 3D face models with mesoscopic details, such as movies, games, and other entertainment industries. With the advancement of this technology, people may conveniently recover high-resolution and high-accuracy 3D face models from multi-view images through specialist devices and efficient algorithms. However, the efficient acquisition of detailed 3D facial models may cause severer privacy violation problems than 2D images. Therefore, we suggest that policymakers should establish an efficient monitoring platform to regulate the illegal spread of 3D face models with private information that may cause ethical problems.

Many images used in this paper are provided by FaceScape dataset\cite{zhu2021facescape, yang2020facescape}, and the corresponding subjects have been known and authorized to use their photos for academic research. We have contacted the author of the FaceScape to confirm that the images in this article are authorized for publication. To protect the identity information of the subjects, we anonymized the face images by mosaicing the eyes. Please note that we used non-mosaic images for training and testing in the experiment, and only mosaic the images in the figures of this paper. 

\section{Acknowledgements}

This work was supported by the NSFC grant 62025108, 62001213, and a gift funding from iQIYI Inc.

{
\bibliography{mybib}
}

\end{document}